%% file: main.tex
\documentclass[runningheads]{llncs}

\usepackage[T1]{fontenc}
\usepackage{graphicx}
\usepackage{csquotes}
\usepackage{booktabs}
\usepackage{todonotes}
\usepackage{tcolorbox}
\usepackage{bbding}

\usepackage[colorlinks=true, linkcolor=blue, citecolor=blue, urlcolor=blue, filecolor=blue, pdfborder={0 0 0}]{hyperref}

\usepackage{color}

\begin{document}

\title{How Well Does AI-Generated Feedback Work? Intrinsic and Extrinsic Evaluation across more than 20,000 EFL Essay Drafts}

\titlerunning{Intrinsic \& Extrinsic Evaluation of WCF}

\author{Steven Coyne\inst{1, 2} \and
Diana Galvan-Sosa\inst{3} \and
Ryan Spring\inst{1} \and
Machi Shimmei\inst{1, 2} \and
Michael Zock\inst{4} \and
Keisuke Sakaguchi\inst{1, 2} \and
Kentaro Inui\inst{5, 1, 2}}

\authorrunning{S. Coyne et al.}
\institute{Tohoku University, Sendai, Japan \\
\email{coyne.steven.charles.q2@dc.tohoku.ac.jp}
\email{\{spring.ryan.edward.c4,machi.shimmei.e6,keisuke.sakaguchi\}@tohoku.ac.jp} \and
RIKEN, Wako, Japan \and
ALTA Institute, Computer Laboratory, University of Cambridge, Cambridge, UK
\email{dg693@cam.ac.uk} \and
CNRS, LIS, Aix-Marseille University, Marseille, France \\
\email{michael.zock@lis-lab.fr} \and
MBZUAI, Abu Dhabi, United Arab Emirates \\
\email{kentaro.inui@mbzuai.ac.ae}}

\maketitle


\begin{abstract}
This study examines feedback in English as a Foreign Language (EFL) writing contexts, focusing on written corrective feedback (WCF).
Large language models (LLMs) can provide WCF at scale, but aligning them with pedagogical best practices remains an ongoing challenge.
WCF meeting criteria like factuality or relevance may still be unsuitable for learning contexts, highlighting the need for extrinsic evaluation based on the learner's perspective.
We deployed WCF systems in a university-level EFL class with nearly 2,000 students, collecting over 20,000 drafts.
We evaluated the generated WCF from two perspectives: intrinsic evaluation by experienced English teachers using a rubric, and extrinsic evaluation via student feedback and engagement metrics.
Results revealed low alignment between teacher expert ratings and student feedback.
These findings suggest that traditional expert evaluation alone may not fully capture WCF's usability or helpfulness from the learner's perspective, highlighting the importance of learner-centered evaluation frameworks for AI-based applications in language education.


\end{abstract}


\section{Introduction}\label{sec:introduction}

Learning a second language (L2) is an iterative, cumulative process, during which errors naturally occur.
In English as a Foreign Language (EFL) writing education, written corrective feedback (WCF) is often provided to help learners notice and correct these errors, increasing their linguistic accuracy~\cite{Brown2023survey}.
However, providing meaningful WCF requires significant time and labor of teachers.
Meanwhile, large language models (LLMs) can generate WCF at scale, but evaluating the educational value of this generated text remains a significant challenge~\cite{maurya-etal-2025-unifying}.

In this study, we explore the evaluation of WCF in EFL writing across two contrasting dimensions: intrinsic aspects (quality as assessed by teachers) and extrinsic aspects (students' interactions with WCF and perceptions of it). 
We report on an ongoing large-scale deployment of WCF systems in a university-level EFL class with 1,999 students, analyzing system performance and student interactions across more than 20,000 essay drafts.
By comparing teacher and student based metrics, we investigate whether expert-approved WCF aligns with what learners find usable or helpful.
We investigate the following research questions:

\begin{itemize}
    \item \textbf{RQ1: Engagement.} Do students use the WCF system? What patterns of engagement emerge across proficiency levels?
    \item \textbf{RQ2: Teacher Perceptions.} How do English teachers perceive the WCF?
    \item \textbf{RQ3: Student Perceptions.} How do students perceive the WCF?
    \item \textbf{RQ4: Evaluation Alignment.} Do teacher perceptions and student perceptions align as measures of WCF quality?
\end{itemize}

\section{Setting and Methods}

This study takes place within a fully online English for General Academic Purposes (EGAP) class, a mandatory course for second-year undergraduate students who scored less than 550 points on a TOEFL ITP test\footnote{\url{https://www.ets.org/toefl/itp.html}}.

All students were are citizens of (anonymized) and share a first language.
The gender distribution is reported as 71.4\% male, 28.6\% female, though some error may exist due to the lack of any additional categories.
The majority of students (n=1701) were of the CEFR framework~\cite{cefr-2001} level B1 based on their TOEFL ITP scores.
A smaller number were A2 (n=59), B2 (n=135), or C1 (n=4).
One hundred students with missing proficiency scores are excluded from analysis.

The class includes seven graded writing assignments (200–250 words) submitted to a homework website, with up to three drafts per essay allowed.
For each submission, students are shown a rule-based automated score that incorporates complexity, accuracy, and fluency (CAF) measures~\cite{kyle_tool_2018}.
After submitting a draft, students can access a ``Feedback'' screen.
A screenshot can be seen in Figure~\ref{fig:feedback_screen}.

The WCF consists of (i) a highlight that contains the language error, and (ii) text that explains the error and what the learner should do.
This is divided into two parts, an ``explanation'' and a ``suggestion,'' following the approach in~\cite{coyne-2025-annotating}.
In our interface, we present these separately as ``What's wrong'' and ``What to do.''
Since the WCF text communicates to the learner about their writing, it constitutes ``metalinguistic'' feedback~\cite{ellis-2008-typology}.
To apply the WCF, students must edit the document themselves, as there is no automatic correction functionality.

\input{figures/feedback_screen}

We use three WCF systems, all of which involve the use of an LLM.
We use the GPT-4o~\cite{openai2024gpt4ocard} model \texttt{gpt-4o-2024-11-20}.
All systems begin with an identical grammatical error correction (GEC) step in which the LLM corrects the draft text and passes ``sentence pairs'' (original and corrected) to subsequent steps.

The \textbf{Template System} is a baseline that does not use LLMs to generate WCF.
Instead, WCF is constructed through string operations using text templates.
These are selected by comparing the sentence pair using ERRANT~\cite{bryant-etal-2017-automatic}, which aligns the two strings and produces ``edits'' annotated with error types, and filling in the template associated with the error type.
This system is inspired by ALLECS~\cite{qorib-etal-2023-allecs}, with the majority of the templates adapted from theirs.

The \textbf{Simple LLM System} uses the LLM to analyze the sentence pair and generate WCF based on the differences between them.
We adapt the ``keyword-free'' prompt from \cite{coyne-2025-annotating}, modifying it so the LLM handles multiple differences between a sentence pair rather than only one.
Additionally, the LLM autonomously identifies and highlights errors rather than relying on an oracle span.
 
The \textbf{Atomic Edit System} adds a step to extract and classify edits, forcing one WCF comment for each.
This is based on the approach in~\cite{song-etal-2024-gee}, in which ``rough edits'' are extracted using difflib and refined into ``atomic edits.''
We use ERRANT to extract the rough edits, as it is more linguistically informed than difflib outputs.
For error classification, we adopt the typology described in~\cite{coyne-2025-annotating}.

For system assignment, we adopt a counterbalanced repeated measures design~\cite{mackey_second_2021} where each student is assigned a different WCF system for each essay assignment, alternating equally between them during the course.

We collect the text of all draft submissions and all generated WCF comments.
When students check the WCF, we log whether each comment was viewed.
We also incorporate a simple student feedback mechanism in the form of thumbs-up (``like'') and thumbs-down (``dislike'') buttons, allowing analysis of student engagement and impressions.
The dislike button permits a text report.

In all, we collected 20,255 drafts from 1,899 participating students after dropping fragmentary data.
Statistics for the full dataset can be seen in Table~\ref{tab:dataset_details}.

\input{tables/dataset_details}

\section{Experiments}

\subsection{Teacher Evaluation of WCF (RQ2)}\label{sec:rating_experiments}

To address \textbf{RQ2}, we hire four teachers to evaluate the generated WCF as part of a manual rating task.
Each had at least nine years of EFL teaching experience.
We adopt the rubric and task guidelines from \cite{coyne-2025-annotating}, except that raters see fully-corrected sentences, rather than isolated corrections.
We also add ``highlight quality'' ratings on a scale of 1-5, and allow raters to flag corrections that are invalid in the first place.
The criteria can be seen in the results report in Table~\ref{tab:teacher_ratings}. 

As our systems generated almost 150,000 WCF comments, we conducted the rating task on a sampled subset.
We set a target of 600 WCF comments, from which we selected 581 based on our sampling procedures.
As we analyze both interaction patterns and system performance, we collect two complementary sets.
The ``Engagement'' set (n=295) comprises WCF that received student feedback via the like or dislike buttons or was viewed, capturing high-engagement contexts and enabling comparison with student perceptions (\textbf{RQ4}).
The ``General'' set (n=286) is a balanced sample across systems and CEFR levels, with inverse square root weighting by error type to mitigate the dominance of common errors, enabling a more robust evaluation of overall system performance (\textbf{RQ2}).

\section{Results}\label{sec:results}

\subsection{Student Engagement and Perceptions (RQ1, RQ3)}

Students can be broadly grouped into those who engaged with WCF and those who ignored it, as 49.76\% never accessed the grammar feedback tab.
Among those who engaged, the average number of WCF comments viewed was 20.4, with some heavy users viewing 10-20 per draft.
The average nonzero views increase with CEFR level among A1 (3.29), B1 (6.57), and B2 (7.98).
For C1, it was 5.75.

Like and dislikes were provided by 259 unique students, of which 71 gave more than one (median of 3).
Ten heavy users gave a median of 18.5 responses.
Details can be seen in Table~\ref{tab:student_rating_results}.
We calculate a student feedback rate (FB Rate) by dividing view count by feedback count.
This is approximately 3\% overall.
The majority of student feedback ratings (92.35\%) are likes.
All text reports referred to the WCF's factual accuracy or highlight quality, as shown in Table~\ref{tab:student_rating_results}.

\input{tables/student_ratings}

\subsection{Teacher Perceptions (RQ2)}\label{sec:results:agreement}
\
\input{tables/teacher_ratings}

Teacher rating task results can be seen in Table \ref{tab:teacher_ratings}.
We use Krippendorff's Alpha to calculate inter-annotator agreement (IAA), finding some limited agreement for correction validity, factuality, and whether the WCF explains what is wrong and why, but low agreement for other criteria.
For ``out of scope'' and ``has what to do,'' raters marked these as insufficient only 0-2 times per batch, and a lack of overlap on these was penalized severely.
Agreement for ``Comprehensible'' increased with the CEFR level displayed, with -0.18 for A2, 0.17 for B1, and 0.39 for B2.
It is \texttt{null} for C1, as all raters selected ``Yes'' for all such cases.

The two LLM-based systems were rated similarly.
The Simple system was rated slightly higher, but a Mann-Whitney U test does not deem it significant (p=0.628).
Both outperformed the Template system.
The Atomic Edit system performed best on highlight quality, which was found significant (p<0.01).

\subsection{Alignment Between Evaluation Types (RQ4)}

We compare the teacher rating results to student feedback on the Engagement set.
While a small difference in mean rating is observed for WCF with likes (4.12), dislikes (3.83), and viewed WCF without student feedback (4.04), it was not deemed significant by a Mann-Whitney U Test (p=0.0625).
Chi-squared tests on binary criteria did not find significant relationships to likes or dislikes.

\section{Discussion and Conclusion}

Some of the low inter-annotator agreement scores may reflect genuine pedagogical disagreement rather than noise, a pattern noted in prior work on subjective pedagogical judgments~\cite{gaudeau-2025-beyond}, and supported here by cases where, during rating task reviews, raters each confirmed their rating and provided a justification.

With mean ratings between 4 and 5, the LLM-generated WCF was largely approved of by the EFL teachers, and students were more likely to assign likes than dislikes.
However, the lack of significant overlap between teacher and student metrics on a per-comment basis suggests that these capture fundamentally different constructs, and
that expert rubric-based ratings alone may not fully capture WCF’s usability or helpfulness from the learner’s perspective.

This study examined LLM-generated WCF for academic writing in a large-scale, real-world EFL course, comparing criteria-based teacher quality ratings with student interaction and feedback analytics.
Our findings reveal that these measures capture distinct, complementary aspects of WCF effectiveness rather than converging.
These findings emphasize the importance of multi-dimensional evaluation of LLM-based applications in language education.

\subsubsection{Limitations}

Our study had several limitations.
There were no paired pre-tests and post-tests, nor experimental groups, so we do not report learning gains.
The student likes and dislikes are sparse signals that are skewed towards high-engagement students.
All students shared the same first language, constraining potential patterns of errors.
The gender ratio was skewed towards male students, and we were not permitted to use gender on a per-student basis during analysis.

\subsubsection{Ethics}

Students agreed to a consent form permitting all analyses in this study.
The participating department handled their personal information and only shared anonymized IDs with the authors.
Use of the OpenAI API was approved by the participating department.
We did not share essay data with any third party.

\begin{credits}
\subsubsection{\ackname} This work was supported by JSPS KAKENHI Grant Numbers JP22H00524 and JP25K03175.
\end{credits}

\subsubsection{Disclosure of Interests}

The authors have no competing interests to declare that are relevant to the content of this article.


\bibliographystyle{style/splncs04}
\bibliography{bib/bibliography}

\end{document}

%% file: figures/feedback_screen.tex
\begin{figure}[!t]
    \centering\includegraphics[width=\columnwidth]{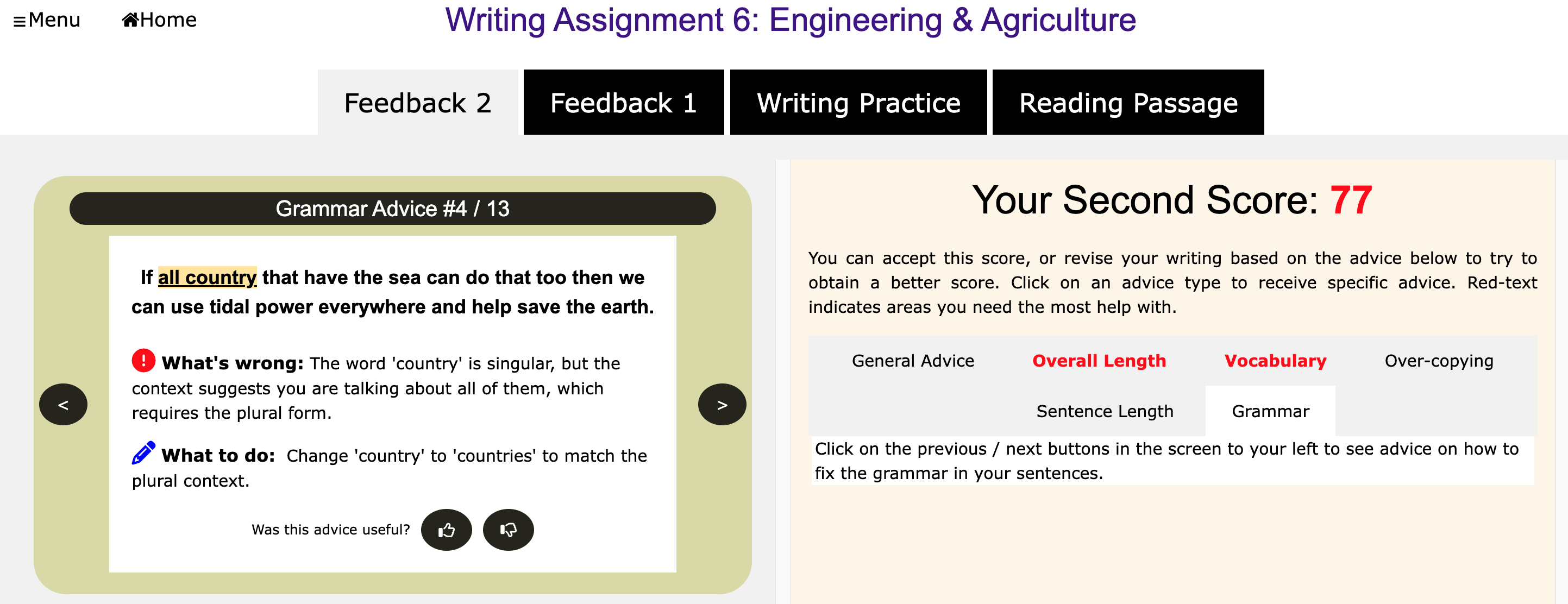}
    \caption{The writing feedback interface, including WCF with like and dislike buttons. The visual interface for WCF is identical for all systems used in our experiments.}
    \label{fig:feedback_screen}
\end{figure}

%% file: tables/dataset_details.tex
\begin{table}[ht]
\centering
\small
\caption{Dataset Details. All values exclude 100 students with fragmentary data.}
\label{tab:dataset_details}
\begin{tabular}{lr p{0.55cm} lr}
\toprule
Number of drafts & 20,255 & & Total WCF count & 144,790 \\
Number of students & 1,899 & & Viewed WCF count & 19,470 \\
Number of unique essays & 11,545 & & Drafts with viewed WCF & 2,927 \\
Essays completed per student & 6.08 & & Views per draft (excluding zeros) & 6.65 \\
Drafts per essay & 1.75 & & Student feedback ratings & 597 \\
\bottomrule
\end{tabular}
\end{table}

%% file: tables/student_ratings.tex
\begin{table}[t]
    \centering
    \small
    \setlength{\tabcolsep}{4pt}
    \caption{Student feedback on WCF comments via the like and dislike buttons.}
    \label{tab:student_rating_results}
    \begin{tabular}{l c cc cc}
        \toprule
         & \textbf{Amount} & \multicolumn{2}{c}{\textbf{Sentiment}} & \multicolumn{2}{c}{\textbf{Dislike Reasons}} \\
         \cmidrule(lr){2-2} \cmidrule(lr){3-4} \cmidrule(lr){5-6}
        \textbf{System} & \textbf{Count (FB Rate)} & \textbf{Likes} & \textbf{Dislikes} & \textbf{Wrong} & \textbf{Highlight}\\
        \midrule
        Template & 261 (3.46\%) & 235 (90.0\%) & 26 (10.0\%) & 12 & 2 \\
        Simple   & 168 (2.83\%) & 160 (95.2\%) & 8 (4.8\%) & 3 & 0 \\
        Atomic   & 159 (2.65\%) & 148 (93.1\%) & 11 (6.9\%) & 5 & 0 \\
        \bottomrule
    \end{tabular}
\end{table}

%% file: tables/teacher_ratings.tex
\begin{table*}[t]
    \centering
    \footnotesize
    \setlength{\tabcolsep}{3.5pt}
    \caption{Manual rating results for each WCF system. Inter-annotator agreement ($\alpha$) is calculated on the full data. ``Invalid corrections'' originate in the shared GEC step.}
    \label{tab:teacher_ratings}
    \begin{tabular}{l c cc c c cc c}
        \toprule
         & \textbf{IAA} & \multicolumn{3}{c}{\textbf{General Set}} & & \multicolumn{3}{c}{\textbf{Engagement Set}} \\
         \cmidrule(lr){3-5} \cmidrule(lr){7-9}
        \textbf{WCF Rating Criterion} & \textbf{($\alpha$)} & \textbf{Tmp.} & \textbf{Smpl.} & \textbf{Atm.} & & \textbf{Tmp.} & \textbf{Smpl.} & \textbf{Atm.} \\
        \midrule
        Correction is valid & 0.70 & 0.92 & 0.98 & 0.94 & & 0.97 & 0.97 & 0.96 \\
        Relevant to error & 0.30 & 0.99 & 0.99 & 1.00 & & 0.99 & 0.97 & 1.00 \\
        Factually correct & 0.65 & 0.92 & 0.96 & 0.92 & & 0.96 & 0.96 & 0.95 \\
        Has what's wrong \& why & 0.61 & 0.28 & 0.99 & 0.98 & & 0.41 & 0.98 & 0.99 \\
        Has what to do & -0.004 & 0.99 & 1.00 & 1.00 & & 0.99 & 0.99 & 1.00 \\
        Is Comprehensible & 0.04 & 0.98 & 0.91 & 0.92 & & 0.98 & 0.90 & 0.97 \\
        Has out-of-scope ($\downarrow$) & 0.22 & 0.01 & 0.006 & 0.005 & & 0.00 & 0.017 & 0.013 \\
        Has direct correction & 0.40 & 0.84 & 1.00 & 1.00 & & 0.84 & 0.98 & 1.00 \\
        Highlight Quality & 0.25 & 4.54 & 4.38 & 4.64 & & 4.65 & 4.41 & 4.65 \\
        Overall Quality & 0.45 & 3.14 & 4.51 & 4.36 & & 3.46 & 4.45 & 4.57 \\
        \bottomrule
    \end{tabular}
\end{table*}

%% file: bib/bibliography.bib
@book{cefr-2001,
    author = {{Council of Europe}},
    title = {Common European Framework of Reference for Languages: Learning, Teaching, Assessment},
    publisher={Cambridge: Cambridge University Press},
    year = 2001,
}

@article{Brown2023survey,
author = {Dan Brown and Qiandi Liu and Reza Norouzian},
title ={Effectiveness of written corrective feedback in developing L2 accuracy: A Bayesian meta-analysis},
journal = {Language Teaching Research},
year = {2023},
doi = {10.1177/13621688221147374}
}

@inproceedings{bryant-etal-2017-automatic,
    title = "Automatic Annotation and Evaluation of Error Types for Grammatical Error Correction",
    author = "Bryant, Christopher  and
      Felice, Mariano  and
      Briscoe, Ted",
    booktitle = "ACL 2017",
    month = jul,
    year = "2017",
    address = "Vancouver, Canada",
    doi = "10.18653/v1/P17-1074",
    pages = "793--805",
}

@inproceedings{song-etal-2024-gee,
    title = "{GEE}! Grammar Error Explanation with Large Language Models",
    author = "Song, Yixiao  and
      Krishna, Kalpesh  and
      Bhatt, Rajesh  and
      Gimpel, Kevin  and
      Iyyer, Mohit",
    booktitle = "Findings of the Association for Computational Linguistics: NAACL 2024",
    month = jun,
    year = "2024",
    address = "Mexico City, Mexico",
    doi = "10.18653/v1/2024.findings-naacl.49",
    pages = "754--781",
}

@article{ellis-2008-typology,
    author = {Ellis, Rod},
    title = "{A typology of written corrective feedback types}",
    journal = {ELT Journal},
    volume = {63},
    number = {2},
    pages = {97-107},
    year = {2008},
    month = {05},
    issn = {0951-0893},
    doi = {10.1093/elt/ccn023}
}

@inproceedings{coyne-2025-annotating,
title="Annotating Errors in English Learners' Written Language Production: Advancing Automated Written Feedback Systems",
author="Coyne, Steven
    and Galvan-Sosa, Diana
    and Spring, Ryan
    and Guerraoui, Cam{\'e}lia
    and Zock, Michael
    and Sakaguchi, Keisuke
    and Inui, Kentaro",
booktitle="AIED 2025",
year="2025",
address="Cham, Switzerland",
pages="292--306",
doi="10.1007/978-3-031-98459-4_21",
isbn="978-3-031-98459-4"
}

@misc{openai2024gpt4ocard,
      title={GPT-4o System Card}, 
      author={OpenAI},
      year={2024},
      eprint={2410.21276},
      archivePrefix={arXiv},
      primaryClass={cs.CL},
      url={https://arxiv.org/abs/2410.21276}, 
}

@inproceedings{qorib-etal-2023-allecs,
    title = "{ALLECS}: A Lightweight Language Error Correction System",
    author = "Qorib, Muhammad Reza  and
      Moon, Geonsik  and
      Ng, Hwee Tou",
    booktitle = "EACL 2023 System Demonstrations",
    month = may,
    year = "2023",
    address = "Dubrovnik, Croatia",
    doi = "10.18653/v1/2023.eacl-demo.32",
    pages = "298--306",
}

@inproceedings{maurya-etal-2025-unifying,
    title = "Unifying {AI} Tutor Evaluation: An Evaluation Taxonomy for Pedagogical Ability Assessment of {LLM}-Powered {AI} Tutors",
    author = "Maurya, Kaushal Kumar  and
      Srivatsa, Kv Aditya  and
      Petukhova, Kseniia  and
      Kochmar, Ekaterina",
    booktitle = "NAACL 2025",
    month = apr,
    year = "2025",
    address = "Albuquerque, New Mexico",
    doi = "10.18653/v1/2025.naacl-long.57",
    pages = "1234--1251",
    ISBN = "979-8-89176-189-6",
}

@article{kyle_tool_2018,
	title = {The tool for the automatic analysis of lexical sophistication ({TAALES}): version 2.0},
	volume = {50},
	issn = {1554-3528},
	shorttitle = {The tool for the automatic analysis of lexical sophistication ({TAALES})},
	doi = {10.3758/s13428-017-0924-4},
	number = {3},
	urldate = {2026-01-27},
	journal = {Behavior Research Methods},
	author = {Kyle, Kristopher and Crossley, Scott and Berger, Cynthia},
	month = jun,
	year = {2018},
	pages = {1030--1046},
}

@inproceedings{gaudeau-2025-beyond,
    title = "Beyond the Gold Standard in Analytic Automated Essay Scoring",
    author = "Gaudeau, Gabrielle",
    booktitle = "ACL 2025 Student Research Workshop",
    month = jul,
    year = "2025",
    address = "Vienna, Austria",
    doi = "10.18653/v1/2025.acl-srw.2",
    pages = "18--39",
    ISBN = "979-8-89176-254-1"
}

@book{mackey_second_2021,
	address = {New York},
	edition = {3},
	title = {Second {Language} {Research}: {Methodology} and {Design}},
	isbn = {978-1-003-18841-4},
	shorttitle = {Second {Language} {Research}},
	doi = {10.4324/9781003188414},
	publisher = {Routledge},
	author = {Mackey, Alison and Gass, Susan M.},
	month = sep,
	year = {2021},
}
